\documentclass{ieeeaccess}
\usepackage{cite}
\usepackage{amsmath,amssymb,amsfonts}
\usepackage{algorithmic}
\usepackage{graphicx}
\usepackage{textcomp}
\usepackage{csquotes}
\usepackage[colorlinks=true,linkcolor=black,anchorcolor=black,citecolor=black,filecolor=black,menucolor=black,runcolor=black,urlcolor=black]{hyperref}
\usepackage{gensymb} 

\def\BibTeX{{\rm B\kern-.05em{\sc i\kern-.025em b}\kern-.08em
    T\kern-.1667em\lower.7ex\hbox{E}\kern-.125emX}}
\begin{document}


\history{Pre-Print Version 1, date of current version 01.06.2022}
\doi{in submission}

\title{RELAY: Robotic EyeLink AnalYsis of the EyeLink 1000 using an Artificial Eye}

\author{\uppercase{Anna-Maria Felßberg}\authorrefmark{1}, \uppercase{Dominykas Strazdas\authorrefmark{2}, 
}}
\address[1]{Institute for Psychology, Otto-von-Guericke University Magdeburg}
\address[2]{Neuro-Information Technology, Otto-von-Guericke University Magdeburg, (e-mail: dominykas.strazdas@ovgu.de)}

\markboth
{Felßberg and Strazdas: RELAY: Robotic EyeLink AnalYsis of the EyeLink 1000 using an Artificial Eye}
{Felßberg and Strazdas: RELAY: Robotic EyeLink AnalYsis of the EyeLink 1000 using an Artificial Eye}

\corresp{Corresponding author: Anna-Maria Felßberg (e-mail: anna-maria.felssberg@med.ovgu.de).}

\begin{abstract}
There is a widespread assumption that the peak velocities of visually guided saccades in the dark are up to 10~\% slower than those made in the light. Studies that questioned the impact of the surrounding brightness conditions, come to differing conclusions, whether they have an influence or not and if so, in which manner. The problem is of a complex nature as the illumination condition itself may not contribute to different measured peak velocities solely but in combination with the estimation of the pupil size due to its deformation during saccades or different gaze positions. Even the measurement technique of video-based eye tracking itself could play a significant role. To investigate this issue, we constructed a stepper motor driven artificial eye with fixed pupil size to mimic human saccades with predetermined peak velocity \& amplitudes under three different brightness conditions with the EyeLink 1000, one of the most common used eye trackers. The aim was to control the pupil and brightness. With our device, an overall good accuracy and precision of the EyeLink 1000 could be confirmed. Furthermore, we could find that there is no artifact for pupil based eye tracking in relation to changing brightness conditions, neither for the pupil size nor for the peak velocities. What we found, was a systematic, small, yet significant change of the measured pupil sizes as a function of different gaze directions.

\end{abstract}


\titlepgskip=-15pt

\maketitle

\section{Introduction}
\label{sec:introduction}
\PARstart{E}{ye} tracking is a well established and popular technology in numerous scientific fields for the examination of various biological (and indirect psychological) parameters regarding eye movements.  Surprisingly, there is little research and no consensus yet about the impact of the surrounding brightness conditions on the measurement of peak velocities (PVs) in saccades. Studies that addressed this issue come to differing conclusions, whether different illuminations have an influence or not. 
\subsection{Eye Tracking}
\label{Tracking}

\Figure[!t]()[width=0.999\linewidth]{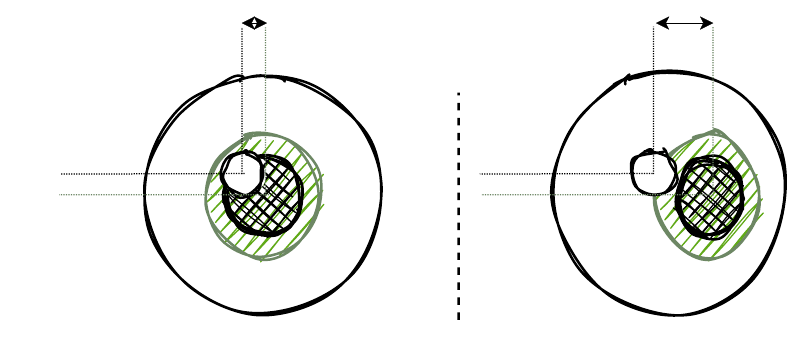}
{Differences in lateral distance between CR and pupil.\label{fig:tracker}}

Throughout history, there have been a lot of methods to investigate the nature of human eyes and their movements: From Electro-Oculography (EOG), Scleral Contact Lens/Search Coil, Photo-OculoGraphy (POG) and Video-OculoGraphy (VOG) to Video-Based Combined Pupil/Corneal Reflection approaches. Today, the latter form the majority of techniques used for eye tracking, as they are fortunately non-invasive and hence rather easy to use. They can track the eyes based on the pupil (P), the corneal reflection (CR) or -- most commonly -- a combination of both (P-CR), which will be described briefly in the following: 

The video-based eye tracker seeks to identify the pupil based on the fact that it appears black and the corneal reflection with the assumption that it appears white. The eye tracker has an infra-red (IR) light source (mostly LEDs) that shines into the subject's eye, resulting in the pupil not reflecting it while the cornea does. Thus, the pupil is estimated with a certain gray scale value that the operator determines in accordance to the surroundings, varying from 0 (black) to 255 (white). The corneal reflection, in video based eye tracking also known as the first Purkinje image, is estimated in the same way \cite{duchowski2017eye}. These two parameters are set in relation to each other. If the subject is looking straight at the eye tracker (which is mostly placed central to a screen), the P- and the CR-signal are close to each other. If the subject is looking away, only the signal of the pupil is moving while the signal from the corneal reflection pretty much stays the same, resulting in differing distances, that become greater, the further away the subject is looking from the tracker. Based on this simple but clever principle, it is possible to determine where a subject is looking. A visualization of this principle can be seen in Figure \ref{fig:tracker}.

While the position of the CR is only slightly affected by the gaze position, the position of the pupil is highly depending on it. If one wishes to track the pupil alone, a rigid fixation for the head is needed in order to prevent its movements. Otherwise, it would not be possible to distinguish between head movements and eye rotation. In the PC-R technique, small head movements can be compensated by subtracting the relative stable CR position signal from the pupil position signal. 

As practical as the measurement using the P-CR technique is due to its obvious advantages, it is also subject to certain susceptibilities to error, as is any given measurement method.

\subsection{The Problem}
Since the measurement technique depends on the signal that the pupil evokes, it is plausible that it can be a severe source of errors. The manufacturer, SR research, recommends for the most applications, that the EyeLink 1000 tracker should be used in centroid mode, which is calculating the position and size of the pupil based on a center of mass approach with the assumption that the pupil is more or less a circle. This tracking algorithm is less vulnerable to noise than an ellipse fitting approach, that may be used if the pupil is significantly occluded \cite{SR_manual}. In the \enquote{centroid mode} the operator can choose to measure the pupil's area or diameter. In their 2011 published article about gaze position influence on pupil size measurement, \cite{gagl2011systematic} explain that it is advisable to work with area instead of diameter if one is interested in not only horizontal but also vertical saccades. The diameter option would be affected stronger for the horizontal saccades while the area option is similarly affected for both directions.

The pupil size is highly depending on a lot of factors, e.g. arousal \cite{bradley2008pupil}, fatigue \cite{yong2004regulating}, intensity of stimuli \cite{bradshaw1967pupil}, cognitive workload \cite{Kahneman1583}, age \cite{guillon2016effects} and -- for this work mostly important -- the illumination of the environment \cite{UBHD-68011840}. As stated by \cite{UBHD-68011840}, a decreasing brightness in the surroundings leads to an increase of the human pupil diameter. The size can range from 1.5 mm up to 8 mm.

\Figure[!t]()[width=0.999\linewidth]{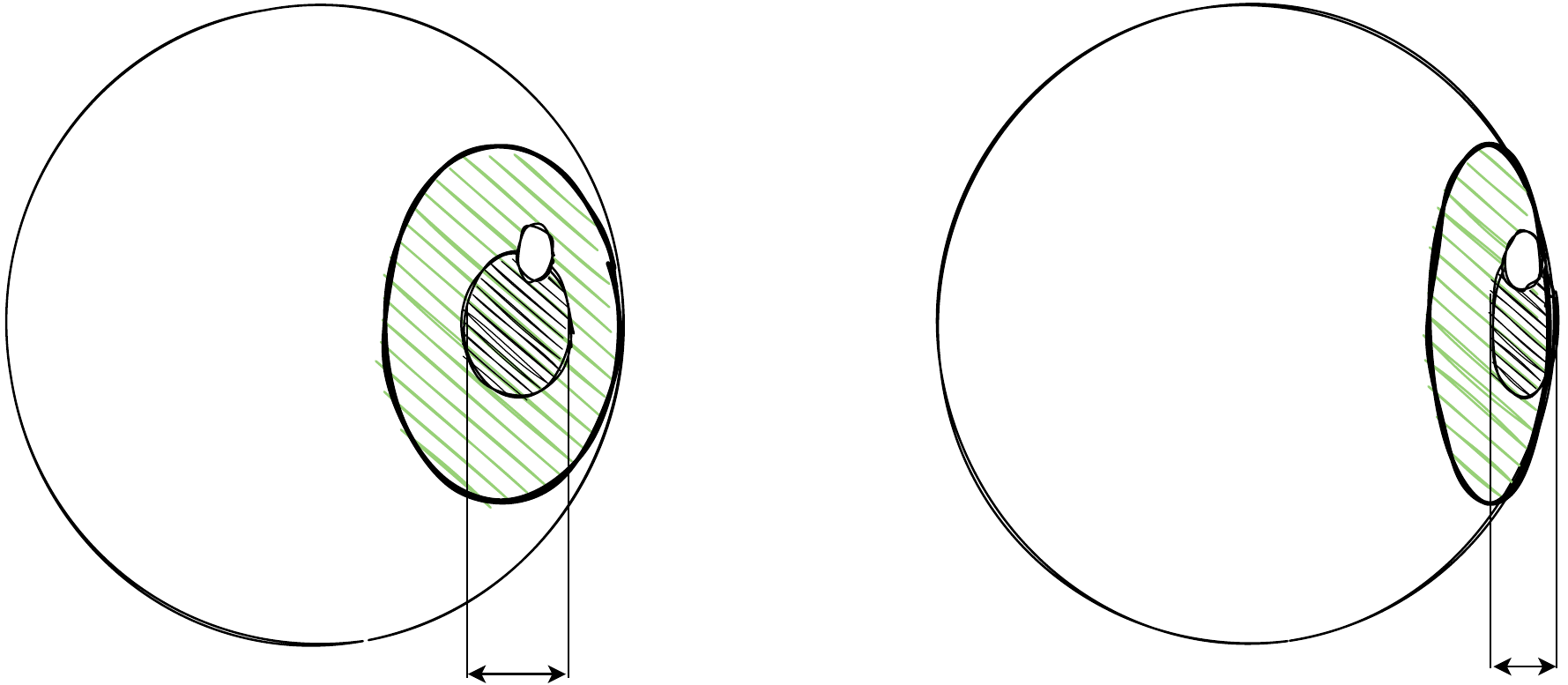}
{The observable shape of the pupil changes with gaze directions. \label{fig:pupil_shape_change}}

The mentioned influences in pupil size are somewhat static in their nature, but there are also impacts of a more dynamic nature that happen during eye movements, such as pupil deformation during motion and changes of the observable pupil shape at different gaze positions. Since the eye consists of elastic material, a deformation of the pupil occurs during saccades as a result of the working forces \cite{nystrom2016pupil}. It is also mostly accompanied by a post-saccadic oscillation (also called saccadic wobble). This deformation, however, influences the pupil's shape and size, which possibly affects the estimation of the pupil as the centroid mode may encounter difficulties estimating something that in theory would be in the shape of a circle but in fact becomes distorted. 

\definecolor{eyegreen}{RGB}{151,208,119}
Furthermore, the gaze position alters the apparent shape of the recorded pupil from a circle to an ellipse, which can lead to difficulties in the estimation of its size and the exact gaze location \cite{ATCHISON200021}, \cite{gagl2011systematic}. An example of the change in shape can be seen in Figure \ref{fig:pupil_shape_change}. A possible contribution to this difficulty could be that the human pupil is not perfectly round, and thus the estimation of its center shifts depending on the perspective that it is looked at.

Another impact on estimation difficulties of the size of the pupil and hence it's position and velocity (as they are all correlated and the latter depends on the others), is a fact that was mentioned by \cite{jay1962effective}: "An additional source for a measurement error might be the anatomy of the iris (more specifically, the thickness of the iris) that defines the pupil". A thicker iris does not cause any trouble while looking straight at the eye, but leads to a stronger effect of apparent shape change of the pupil when looked diagonally at it, as it would have been if one looks at a tube. The seen ellipse shape becomes even narrower when looking at the hole (pupil) from the side when the iris is thick than it would be for a slim iris \cite{gagl2011systematic}. Figure \ref{fig:thicc} depicts this effect.

\Figure[!hb]()[width=0.999\linewidth]{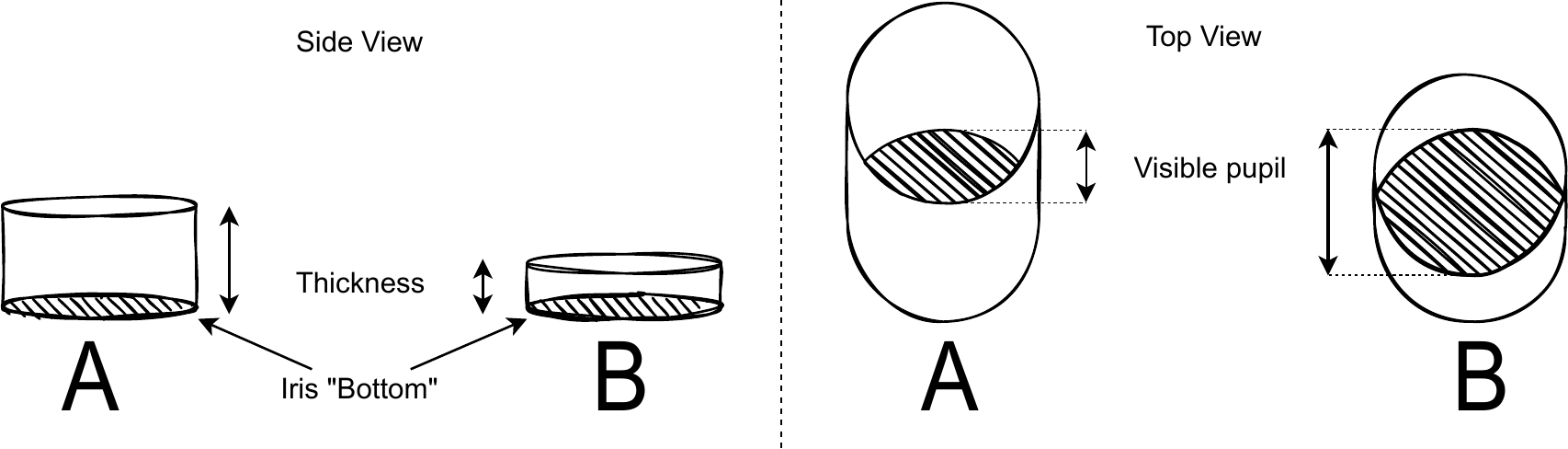}
{Different visible pupil sizes depending on the thickness of the iris. \label{fig:thicc}}

\newpage
\subsection{Relevant works}
The estimation of the pupil is relevant to draw conclusions not only about gaze positions but peak velocities as well, as they are dependent on the first. It is important to take these impact factors into concern. Another obvious impact on the pupil size and thus probably the measured peak velocities that should be considered, is the illumination of the surroundings. But despite the logic relation that they  share, there is little research about the consequences of illumination on peak velocities. 

There are some statements claiming peak velocities in the dark would be slower than in the light \cite{craighead2002corsini,Walker2012}. But often they rely on old studies that either compared memory guided saccades (MGS) in the dark towards visually guided saccades (VGS) in the light \cite{Henrikssonetal.1980} or did not even claim said statement \cite{Bahill1975,BeckerFuchs1969}. But memory guided saccades have lower peak velocities than visually guided saccades \cite{SMIT1987}. This finding could be confirmed by \cite{Felberg2018TheEO}. They found that MGS were significantly slower than VGS, regardless of the surrounding brightness conditions. Furthermore, they found a non-linear change of peak velocities and durations in saccades made under three different brightness conditions, with the dark condition being the slowest, medium the fastest and bright in between them. Even when the variance, explained by pupil size changes, was regressed out, the data still showed lower peak velocities for saccades in the dark and the light than in the medium brightness condition, thus remaining the same non-linear roof shape. 

It is the subject of speculation, what could be the source of these findings. Either it is something that is produced by the factor human or the eye tracker in varying lightning conditions. The study was conducted using an EyeLink 1000 in Pupil-Corneal Reflection (P-CR) mode. The "roof-shaped" pattern of the peak velocities can be observed in Figure \ref{fig:roof}. 

\Figure[hb]()[width=0.999\linewidth]{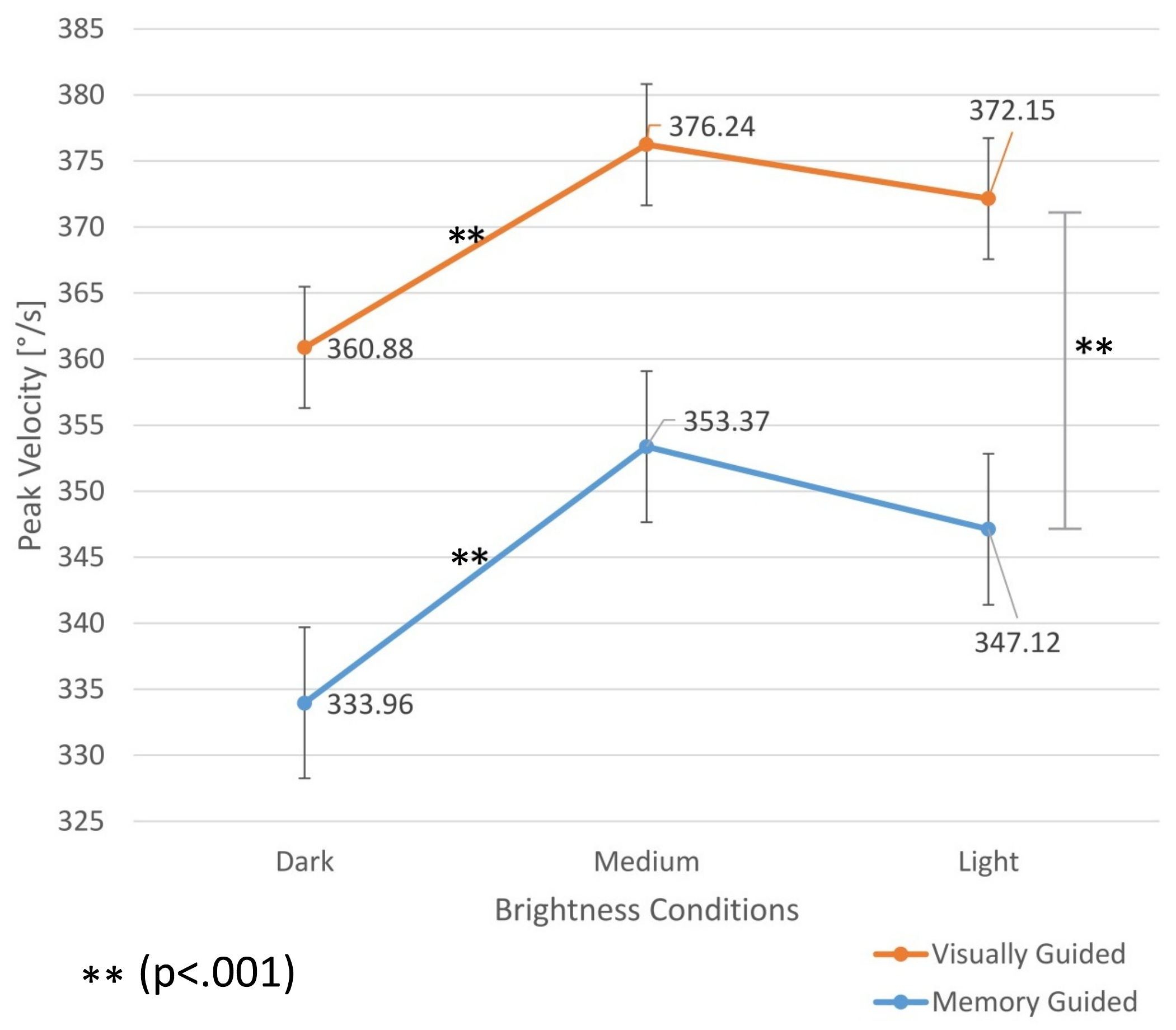}  
{Mean peak velocities over all participants for an amplitude of 10°, from the study of \cite{Felberg2018TheEO}.\label{fig:roof}} 

In 2016, \cite{nystrom2016pupil} reported similar results when they measured the peak velocities of four participants under seven different brightness conditions. They even found the brightest condition (which led to the smallest pupil size) to produce the lowest peak velocities and the second-darkest condition the highest. The darkest condition again lead to lower peak velocities. This was true for all measured signals that they used: CR, pupil and gaze position, though to different extents. As a consequence, they also found a non-linear shape as a function of brightness. They interpreted the change of peak velocities in the gaze signals to be a function of the different pupil sizes. It is noteworthy that the changes in pupil size, found by \cite{nystrom2016pupil}, were of a linear nature, but the changes in peak velocities were not. They admitted themselves, that the first therefore cannot be the only explanation for the found phenomenon. 

In a recent work, \cite{hooge2021pupil} researched what they call the Pupil Size Artifact (PSA), which is an apparent gaze deviation that occurs, when the pupil changes its size, although the eyeball does not rotate (for instance, during fixations). This was also denoted by \cite{wyatt1995form}, who stated that when pupil size changes, the center of the pupil can shift, which affects the gaze position estimation, as the eye tracker seeks to determine the position based on it.

These different studies show that the results and therefore the agreement on the metrics of saccadic peak velocities in dependency of surrounding illumination and pupil sizes is not given to this day. There is no consensus regarding the influence of brightness conditions. It becomes evident that it is a very delicate topic to try to entangle the multiple influence factors on something that is in theory such an easy principle.

While it is true that the pupil size indeed can be a factor, we will ask here how it is the other way around? To what extent could the gaze position have an influence on the pupil size and therefore the measured peak velocities? To deal with this question, we aimed to eliminate the influence caused by the pupil size by utilizing an artificial eye with a constant pupil size, in a robotic motion system.

Furthermore, we controlled the velocity of the robotic eye's movements and the surrounding illumination. As a result, we can not only draw conclusions about the impact of the gaze position but also about the brightness conditions regarding the possibility, that it could lead to an artifact in the eye trackers pupil and thus gaze and velocity estimation. We used the EyeLink 1000 by SR research, as it is one of the most common used, state-of-the-art systems. 

Our constructed system, which will be described in the subsequent section, is the \textbf{R}obotic \textbf{E}ye\textbf{L}ink \textbf{A}nal\textbf{Y}sis apparatus, henceforth referred to as RELAY.

The remaining paper is organized as follows: Section~\ref{sec:Method} describes the RELAY apparatus and our proposed methodology and experiments. The results for accuracy and precision and artificial saccades experiment are presented in Section~\ref{sec:results}. Finally, Section~\ref{sec:conclusion} concludes this paper.

\section{Method}\label{sec:Method}

\subsection{RELAY}

The idea is to use a mobile, lightweight, precise and programmable apparatus with an artificial eye to mimic representative, human-like saccades. The mechanical build is based on two stepper motors, one for the X-axis and one for the Y-axis of the eye, connected to an analog stick (dual axis hinge) from a game pad, that allows an axis unrestricted movement. The analog stick also contains built-in potentiometers for each axis, necessary for initial calibration.

The stepper motors, typically found in 3D printing or CNC applications, were paired with an Arduino board and a stepper driver expansion shield and driven in 16-micro-stepping-mode, resulting in a smooth and quiet motion and a theoretical addressable resolution of $0.1125\degree$ per micro-step. The positioning error, considering the potentiometers, Arduino analog-digital conversion, the used stepper motors and drivers, can be estimated as $\pm2$ micro-steps or $\pm0.225\degree$ not including backlash or mechanical imperfections. A detailed assembly information, parts list and Arduino code is published as an open source git repository \cite{dominykas_strazdas_2022_6591910}. The final RELAY assembly can be seen in Figure \ref{fig:RELAY_Front}. 


\Figure[!b]()[width=0.999\linewidth]{Graphics/fig_5_front.JPG}
{The final RELAY assembly: two stepper motors connected to an analog stick with an artificial prop eye. An Arduino UNO in the back with a CNC expansion shield and two stepper drivers power the motors (12v). The lower motor is attached to the base using a dual tilt adjustment adapter. \label{fig:RELAY_Front}}

\subsection{Experimental Setup}
The experiment was run in the eye tracking laboratory of the Otto-von-Guericke-University in Magdeburg, Germany. The build-up with the eye tracker and the participant monitor were enclosed in a sound-proof cabin, thus leaving the surrounding very well controlled in behalf of the illumination conditions because it was possible to set the display illumination as the only light source. The host-PC (that runs the eye tracker) and a laptop from where the experiment was controlled, were placed in the laboratory outside the cabin. A schematic overview containing the connections between the different components can be seen in the Figure \ref{fig:setup0}. The laptop (\textit{2,60 GHz, 16 GB RAM, 1 TB SSD, Windows 10}) on which the experiment code was run, using the Psychophysics toolbox version 3 \cite{ThePsychophysicsToolbox} for MATLAB\raisebox{1ex}{\tiny{\textregistered}} (Mathworks, Natick, MA, v. R2020b), was connected to the host-PC and the display-PC. 
It was further connected with the Arduino control unit, on which a program written in C++ was run to control the movements and a PlayStation\raisebox{1ex}{\tiny{\textregistered}} 4 controller to achieve better ergonomics for starting different RELAY modes corresponding to a button press. The scripts and program code used in the experiment can be found here \cite{dominykas_strazdas_2022_6591910, Strazdas_Felsberg_2022}.

\Figure[!b]()[width=0.92\linewidth]{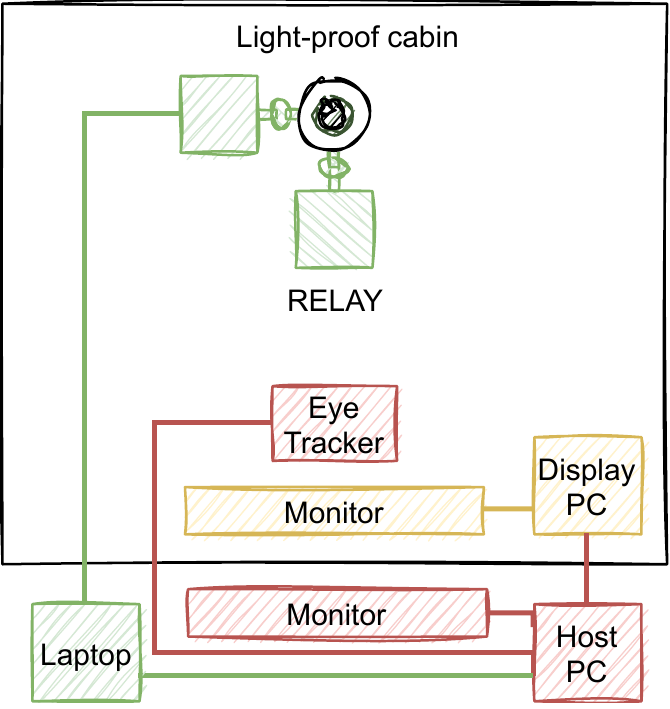}
{Schematic overview of the test setup. \label{fig:setup0}}

Two iiyama ProLite GB2488HSU-B1 (EOL) 24 inch (display area: $298.89~mm$ x $531.36~mm$) monitors with a refresh rate of 144Hz served as display- and "participant" PCs. The monitor's resolution was 1920 x 1080 pixels. The distance from the top knob of the eye tracker to the chin rest was $530~mm$. The bottom knob of the eye tracker was centered horizontally along the monitor. In accordance to the statements of \cite{gagl2011systematic}, we used the Eye tracker in \enquote{centroid mode} with the pupil area option to ensure, that the signal noise was low for the used gaze directions, making the results comparable to other studies which mainly use these options. The used sampling rate of the eye tracker was 1000 Hz.

The RELAY system was clamped to the chin-rest and positioned in such a way, that the artificial eye would approximately be in the same position, compared to a human participant. The chin-rest height slider and the RELAY tilting adapters were then used to bring the apparatus to a realistic position, right in front of the monitor.  The view of the cabin and the experimental set-up can be seen in Figure \ref{fig:setup1}.

The distance from the artificial eye to the screen was set to $927~mm$, according to the recommended distance of at least 1.75x the width of the display area by the manufacturer \cite{SR_manual}. This was done to fulfill the convention as it was strictly speaking not necessary for this experiment as the artificial eye is "blind" and thus does not see the monitor. But it was important to get the right screen coordinates in pixels for the 13-point calibration grid that was applied before testing and to be able to get saccades of the desired amplitudes. The required coordinates for the calibration grid were calculated with the SR research calibration-coordinate-calculator \cite{SR_research_2020_sr}. 

\Figure[!ht]()[width=0.999\linewidth]{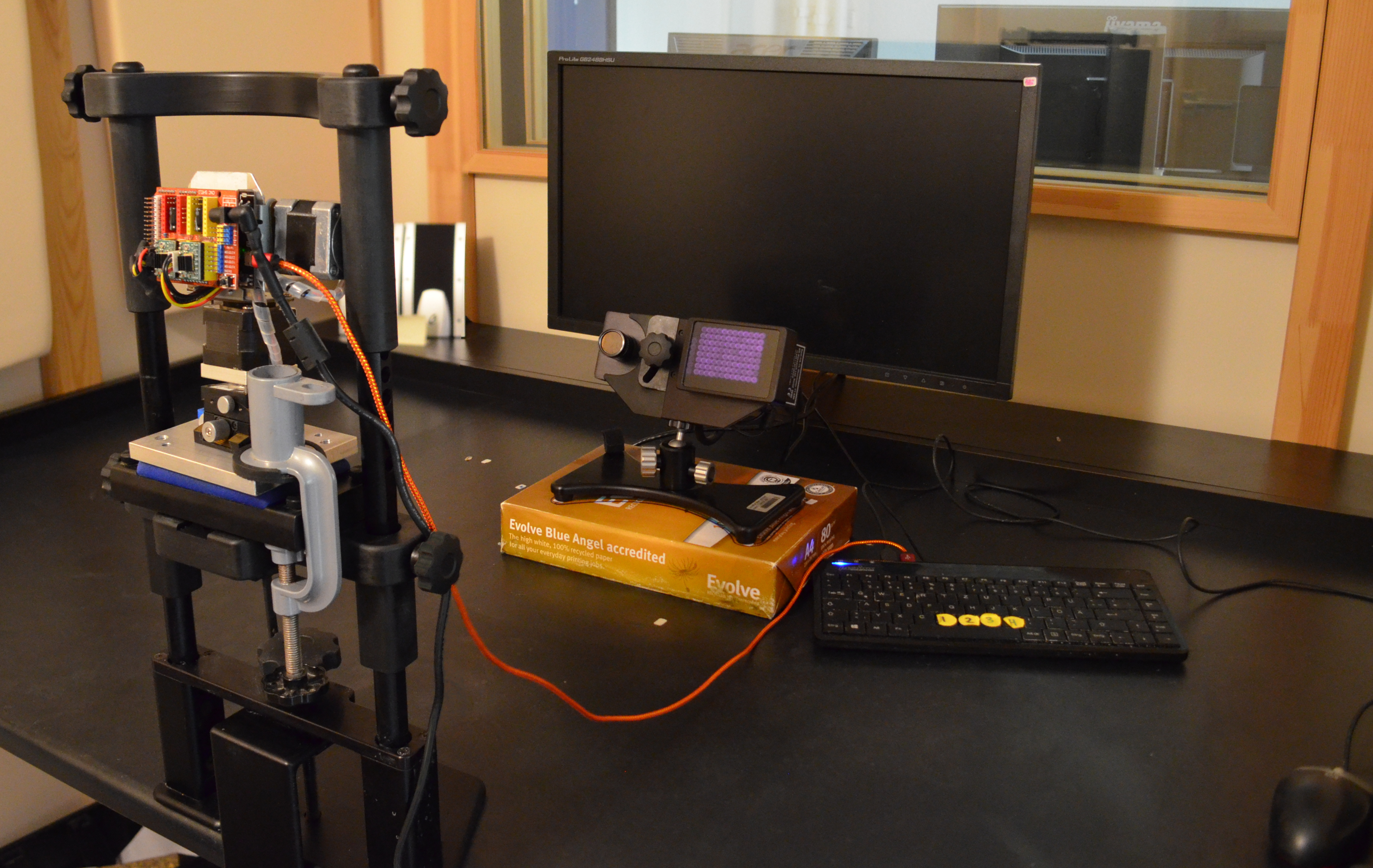}
{Setup in the sound-proof cabin. The operator sits on the other side of the glass window which was occluded for the experiment.
\label{fig:setup1}}

\subsection{Accuracy and precision experiment}

A key element for the success of the experiment is that the RELAY has to move precisely and accurate. Precision, in a narrower sense, means that on multiple measurements with identical setting, the results should also be close to each other (ideally the same). The more are the results alike, the better. Accuracy is given, when a measurement leads to results that are very close (ideally exactly) to the true value. For our moving artificial eye this means, that when RELAY is moving frequently towards a predefined position, it should always move in the same amount, resulting in the same (or close) values for the position, to be precise. Furthermore, when the real position matches the desired position, then it is also accurate. If the movements are not precise, there is no chance to control the recorded values by the eye tracker at all. The input (predefined movements) must be reliable to attain a reliable output (measured movements). If the precision is known and sufficient, everything else in the system can be adapted in order to become accurate as well. This is the base that the whole system can be calibrated and validated by the eye tracker. Thus, it had to be tested first. To ensure an adequate comparability with human saccades and still have good controllable, smooth kinematics, the speed of RELAY's movements had to be determined. The motor speed, acceleration and deceleration are controlled by the driver by increasing the pulses per second (pps) which correspond to a micro-step until the desired speed and position are reached. Trials with a speed of 100, 1000, 3200 and 5000 pps were run. For the acceleration a value of 80\,000 pulses per second² was set. With 100 and 1000 pps being too slow for the eye tracker to correctly detect saccades (they were interpreted as micro saccades) and 5000 being too fast, as it resulted in problematic overshootings out of the monitor screen and thus the trackable range, 3200 pps were chosen. Converted into degrees per second, it results to 360°/s. \
7 patterns were chosen for the accuracy and precision experiment: a horizontal and a vertical line (consisting of several saccades of differing length along the x-, respective the y-axis), multiple horizontal and multiple vertical lines along the screen, a diagonal cross ("X"), a 13-point calibration grid used to calibrate the eye-tracker and a complex pattern consisting of 160 random saccades. After recording the different patterns with the eye tracker, a second run was conducted using a laser dot and a mirror. For this, a canvas with a laser diode in its center was placed in front of the eye. The laser aimed directly at the middle of the artificial eye. The eye was equipped with a little front surface mirror that reflected the laser back onto the canvas depending on the gaze position of the eye. It was adjusted, so it would reflect the laser back to the source in the neutral position. A schematic depiction of the function principal of the laser measurements setup can be seen in Figure \ref{fig:laserprec}. The patterns drawn by the laser depict the viewing direction of RELAY in an enhanced manner, as the angle of the laser reflection is double the angle of the viewing direction of the eye.
\Figure[!hb]()[width=0.999\linewidth]{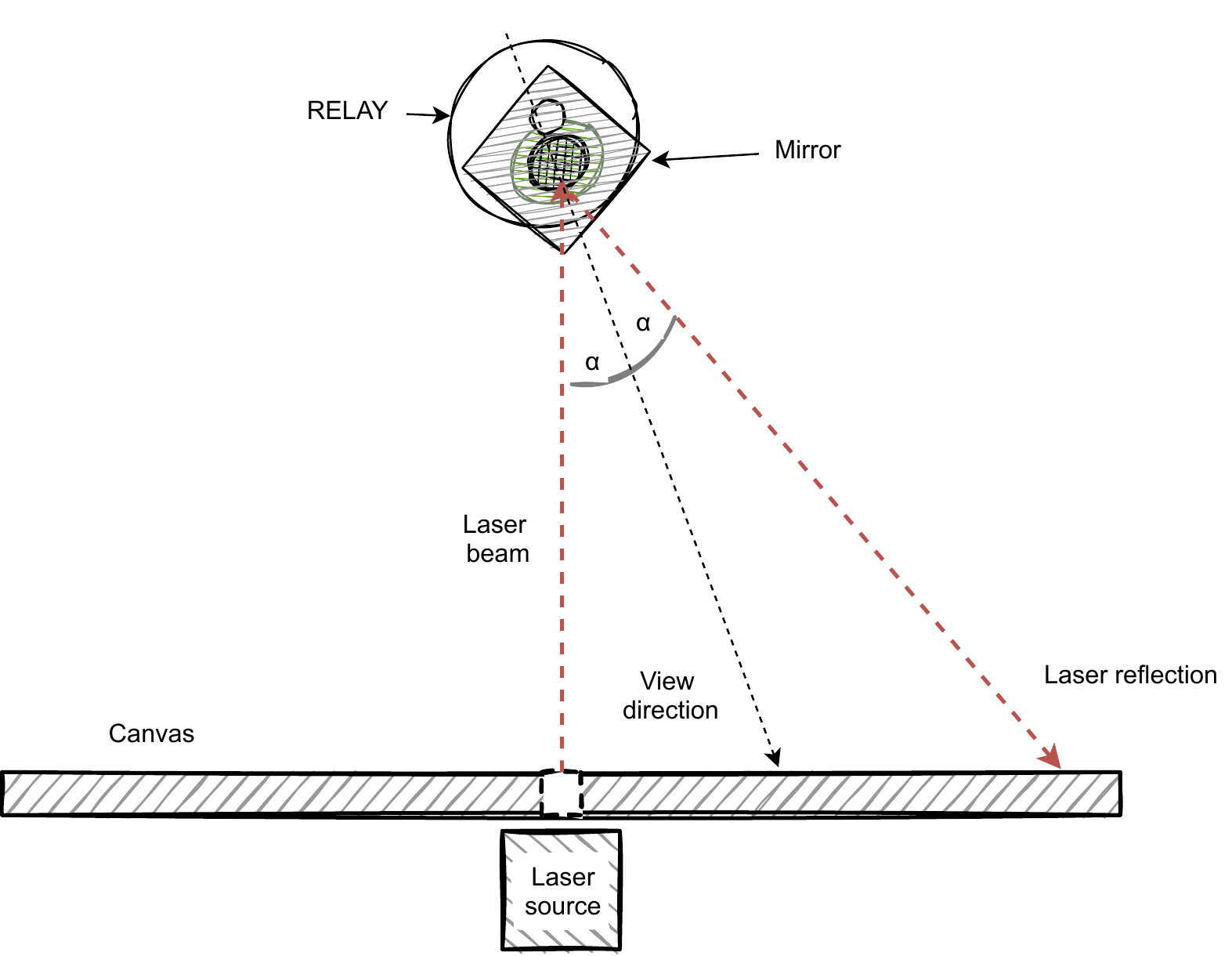}
{Laser beam reflected by the mirror on the eye.\label{fig:laserprec}}

This is caused by the law of reflection for this case of specular reflection, where the reflected ray of light, here the laser pattern on the canvas, emerges from the reflecting surface at the same angle to the surface normal as the incident ray, namely the laser input. The surface normal in this case is the mirror attached perpendicular to the viewing direction. The resulting patterns on the canvas capture the movements of the artificial eye with great detail (more than twice the resolution) due to the trigonometric relations between the canvas, laser and the mirror. With a Panasonic Lumix DC-FZ83, Pictures were taken in long-time exposure mode, making the patterns visible. The Figures \ref{fig:laser_set1} and \ref{fig:laser_set2} show the laser measurement setup. 
With this procedure, it was possible to compare the planned patterns to the actual patterns that were done by the robot and the patterns that the eye tracker recorded. The pictures from this procedure are referred to as laser pictures and can be seen in the Results section.
\Figure[!h]()[width=0.999\linewidth]{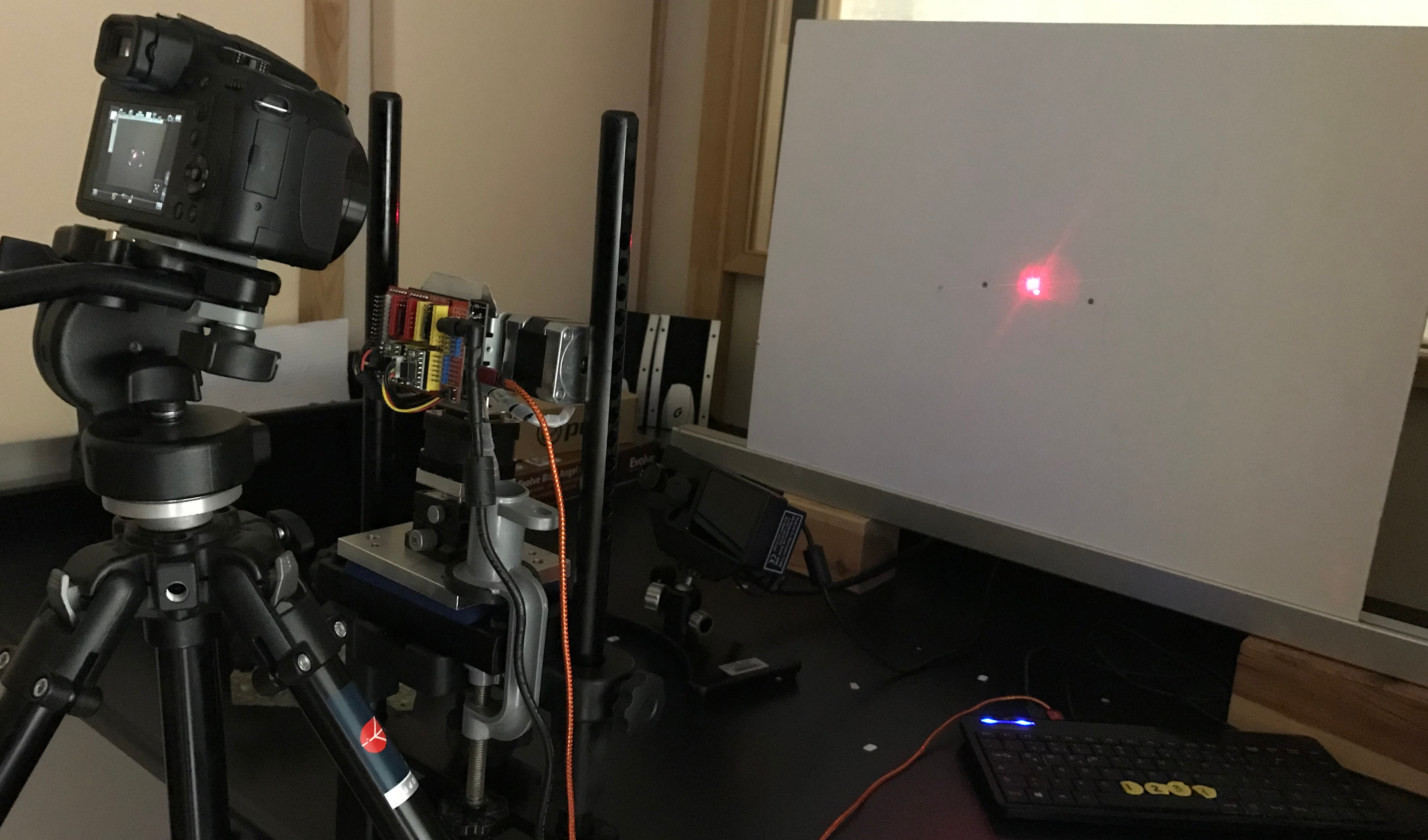}
{The experiment setup with laser and canvas.\label{fig:laser_set1}}
\Figure[!h]()[width=0.999\linewidth]{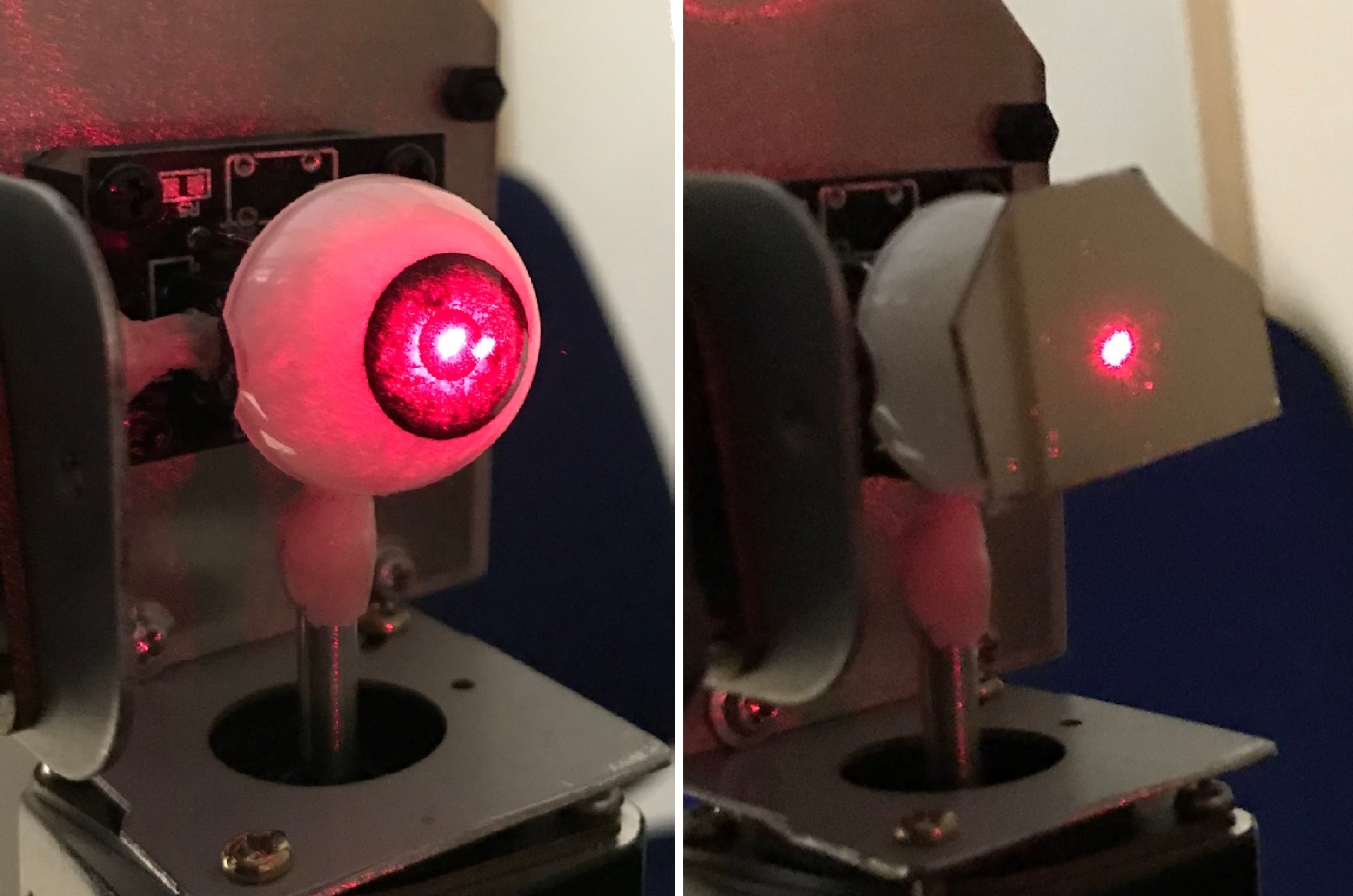}
{The artificial eye and the mirror.\label{fig:laser_set2}}

For the precision measurement, RELAY performed 100 trials of the single horizontal line, the single vertical line and the calibration grid in behalf of comparing the raw data sets and the calculated saccades for each pattern. 

\subsection{Artificial saccades Experiment}
The experimental run was following the design of the initial study by \cite{Felberg2018TheEO}, but only the VGS. This means that the saccades by RELAY were done with the same three different gray scales for the background and the targets on the screen as before, leading to similar brightness conditions (\textit{dark, medium, light}). As the measurement took part in a new laboratory room with another monitor in the cabin, the luminance and contrast measurement had to be repeated. The luminance of the display was measured on five positions of the display (all corners and the center). From these values, the mean was calculated. 

Together with the measured luminance values for the target dots, contrasts were calculated. Although the identical gray scales were used, it led to deviating values (0.68, 0.83 and 0.85) than in the previous experiment due to the other used display. Nonetheless, it was decided to use the same brightness values in order to make them comparable. The conduction was done with the identical eye tracker that has been used before.

\Figure[!b]()[width=0.999\linewidth]{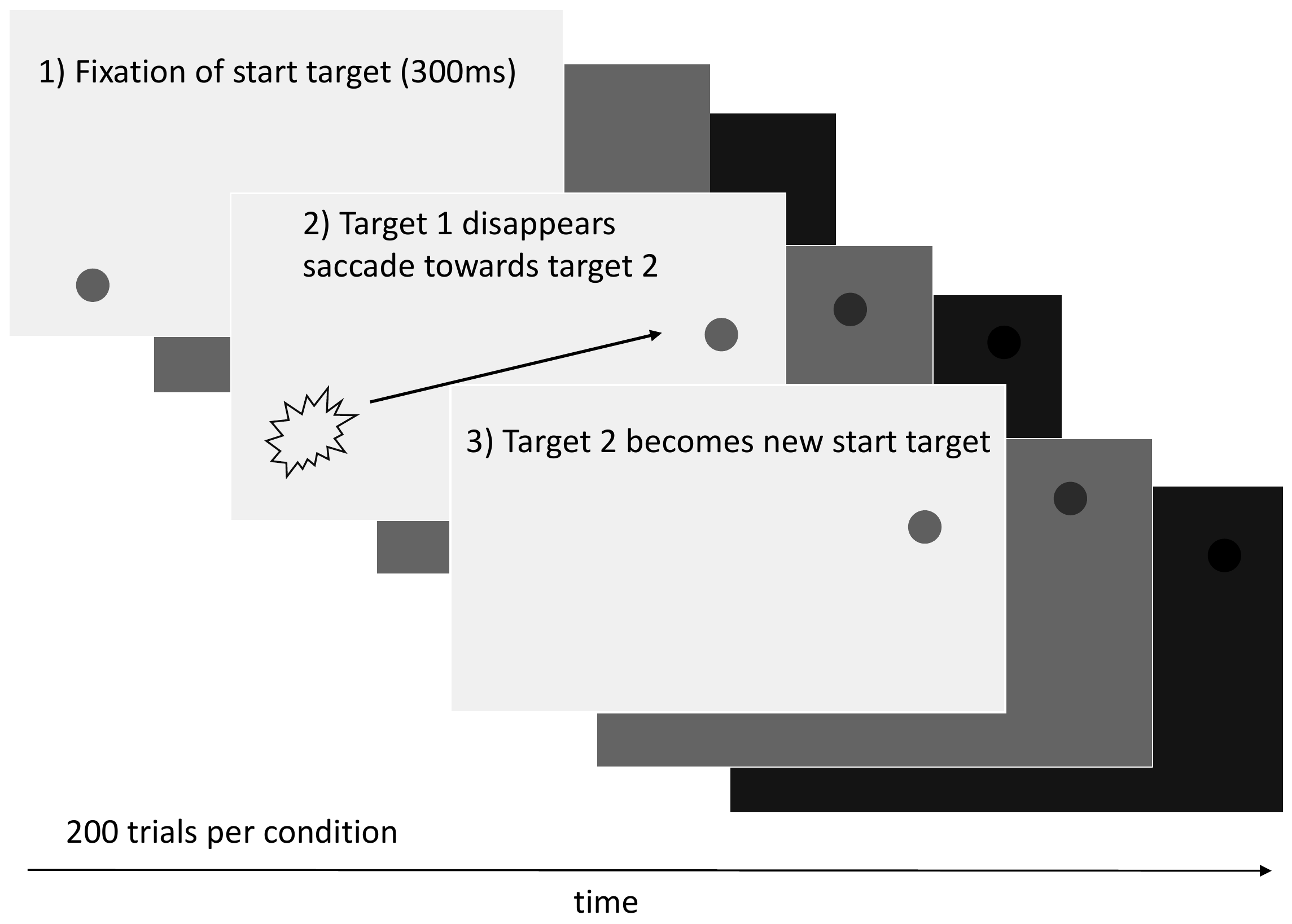}
{Visualization of the trial design for all three brightness conditions. \label{fig:design}}

The experiment was designed to mimic 200 participants doing 200 saccades per condition, resulting in a total of 120,000 saccades (40,000 per condition). When the artificial eye was looking at a target dot for 300 ms (fixating), the dot vanished while the next target dot appeared on another position on the screen. This was followed by a saccade to the second dot (with 3200 pps), which became the new start point for the upcoming saccade after another fixation for 300 ms and so on. After initial calibration and validation to the 13-point calibration grid, the experiment was running for several hours without any interruptions, so that the calibration only had to be done once. 

The positions of the dots were calculated randomly for every \textit{participant}, ranging from 4.5° to 14.1° of visual angle. The distances of the target points were chosen to correspond to the possible range minus $10\%$, so that if overshoots occurred, the saccades would not protrude from the screen. Over the three conditions, the positions stayed the same for every participant in order to make them comparable to each other. Figure \ref{fig:design} explains the sequence which was also used in the prior experiment with real human subjects.

\section{Results}\label{sec:results}

\subsection{Accuracy: Laser Patterns}
The distance from RELAY to the canvas was $70~cm$ when the long time exposure pictures were taken. For the simple patterns, the distances between every position (seen as red dots in Figure \ref{fig:laserneu1}) was set to 8 micro-steps, giving 0.9° each. For example: in the case of the single horizontal line, it summed up to 15 steps to the left and 15 steps to the right. This corresponds to theoretically 27° (or $33.6~cm$) in both directions for the outermost coordinate (note that the angle is doubled because of the mirror on the artificial eye, as explained in the Methods section). 

With the dimensions of the canvas known to be $67.9~cm\times~49.8~cm$, it is possible to calculate the real driven angle by the robot. The rightmost point is directly at the edge of the canvas, i.e. exactly $33.95~cm$ from the center where the laser input is. Hence, the deviation of the theoretical and the real coordinate is $0.35~cm$, respective, 0.29° of visual angle.

When drawing conclusions about the accuracy based on the taken pictures, it is important to notice that the different angles of the camera and the eye towards the canvas (parallax), as it could only be put somewhere close to the eye but not on the exact same spot, lead to a small distortion of the taken pictures. Another distortion, a so-called barrel distortion, originated from the objective lens of the camera. To compensate for this effect, an image registration with perspective correction was done with the picture editing software Gimp 2.10.22 \cite{gimp}. 
\Figure[!b]()[width=0.999\linewidth]{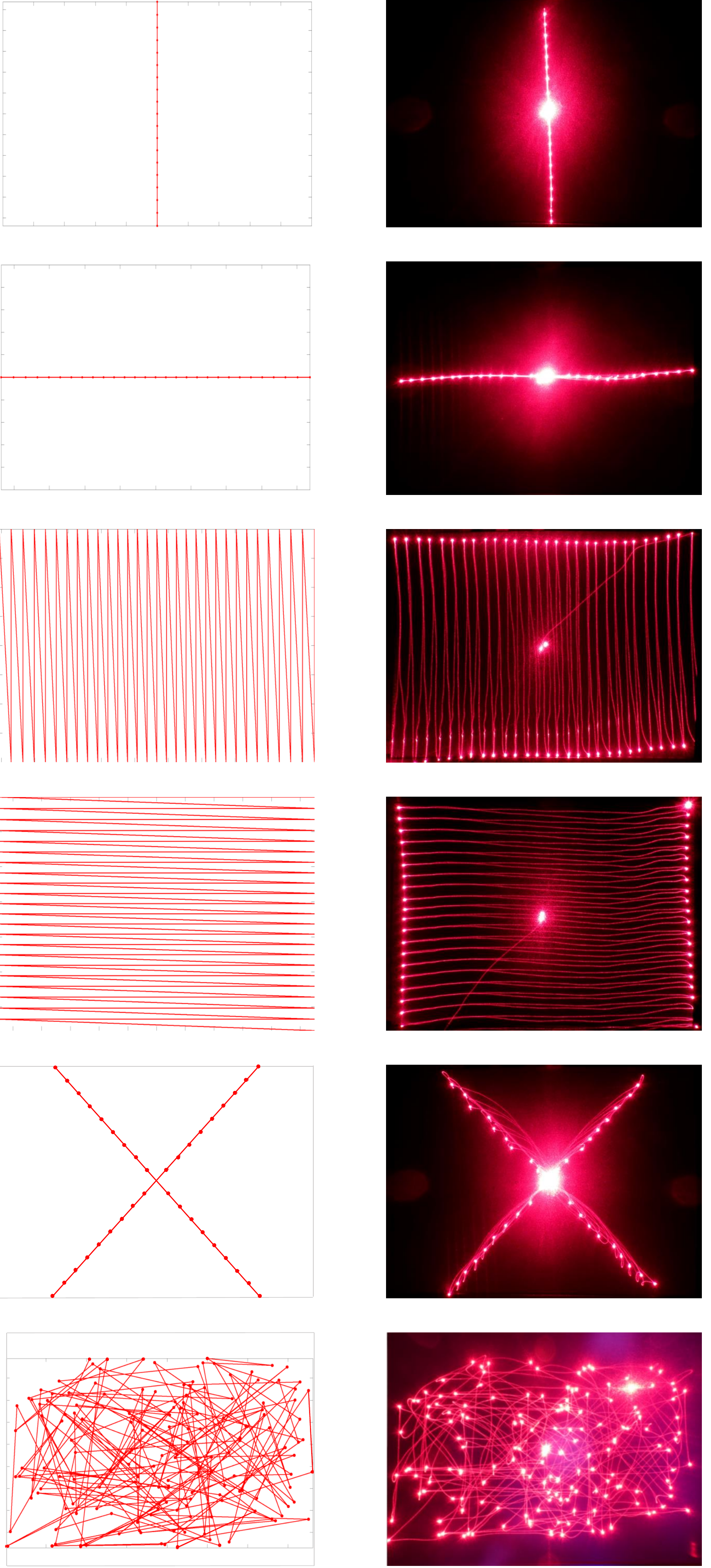}
{Planned patterns vs. long exposure laser patterns. \label{fig:laserneu1}}

In Figure \ref{fig:laserneu1} a comparison of the planned patterns and the corresponding laser pictures can be seen. The planned patterns show the simulated resulting reflection based on the trigonometric values for the distance, canvas size and the original functions, as used to program RELAY.

\subsection{Accuracy: Calibration Pattern}
The accuracy of the planned 13-point calibration pattern vs. its raw data from the eye tracker was determined by calculating the absolute deviation from the mean of the measured X- and Y-positions, during fixation periods, over all 100 trials ($\overline{X}$) to the planned, \textit{true} coordinates ($\tau$). With these it was possible to determine the absolute deviations in degrees of visual angle and thus, the accuracy. The values can be seen in Table \ref{tbl:accuracy}. 
The resulting deviations of the 100 trials for the calibration grid were ranging from -13.3~px to +38.5~px (or -0.23° to +0.66°) for the X-axis and from -18.5~px to +17.4~px (or -0.32° to 0.30°) for the Y-axis. The corresponding laser picture with the planned, and the real pattern, can also be seen in the upper panel of Figure \ref{fig:13point}. \\

\begin{table}
  \caption{Accuracy for the 13-point calibration grid}
  \label{tbl:accuracy}
  \includegraphics[width=0.999\linewidth]{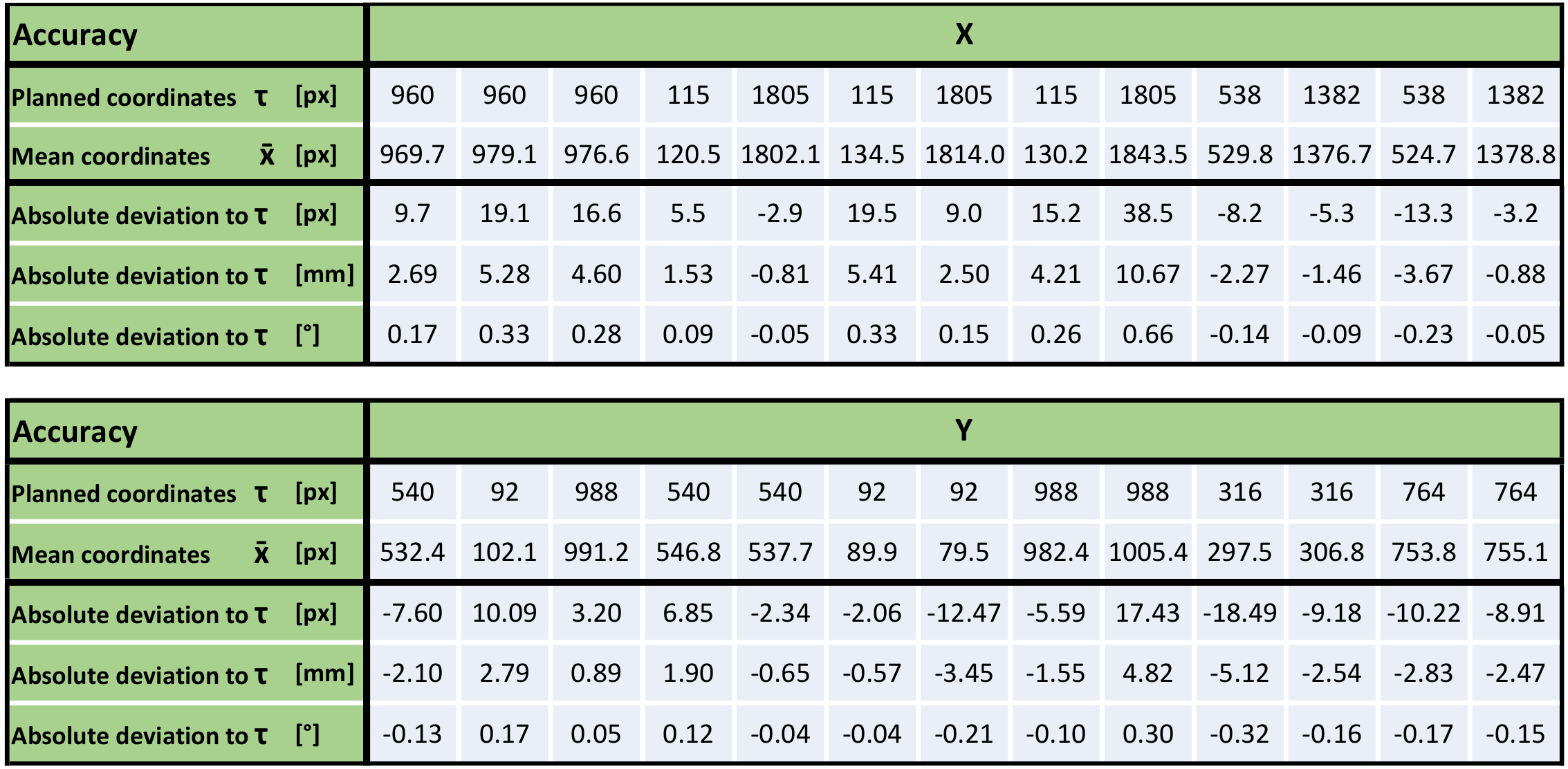}
\end{table}
\newpage
\subsection{Precision}
The first step of precision analysis was done with the usage of the software, Data Viewer, Version 4.1.211, by SR Research. This program is able to show the trial data either in a 2D temporal graph mode or in a spatial overlay view, based on the recorded coordinates. It makes it possible, to see all recorded gaze positions and the movements between them of a single trial at once (Note: in this case, one calibration pattern is viewed as one trial). 

The spatial overlay view of the program can be used with the recorded raw sample data or the performed saccades, which are based on a calculation from the first, done by the program. Unfortunately, we can only speculate about the exact calculation method for the saccades, used by that program. Furthermore, the positions and durations of fixations are observable. This can come in handy, to provide a good overview on the data. 

We used the SR Research Data Viewer to create an accumulated overlay of the calculated saccades from the 100 precision trials for the 13-point calibration grid and the simpler horizontal and vertical lines. Sadly, there was no option to do the same with the raw sample data sets, so they had to be extracted for a manual overlay. This was again carried out with the use of Gimp 2.10.22 \cite{gimp}.
\Figure[!b]()[width=0.999\linewidth]{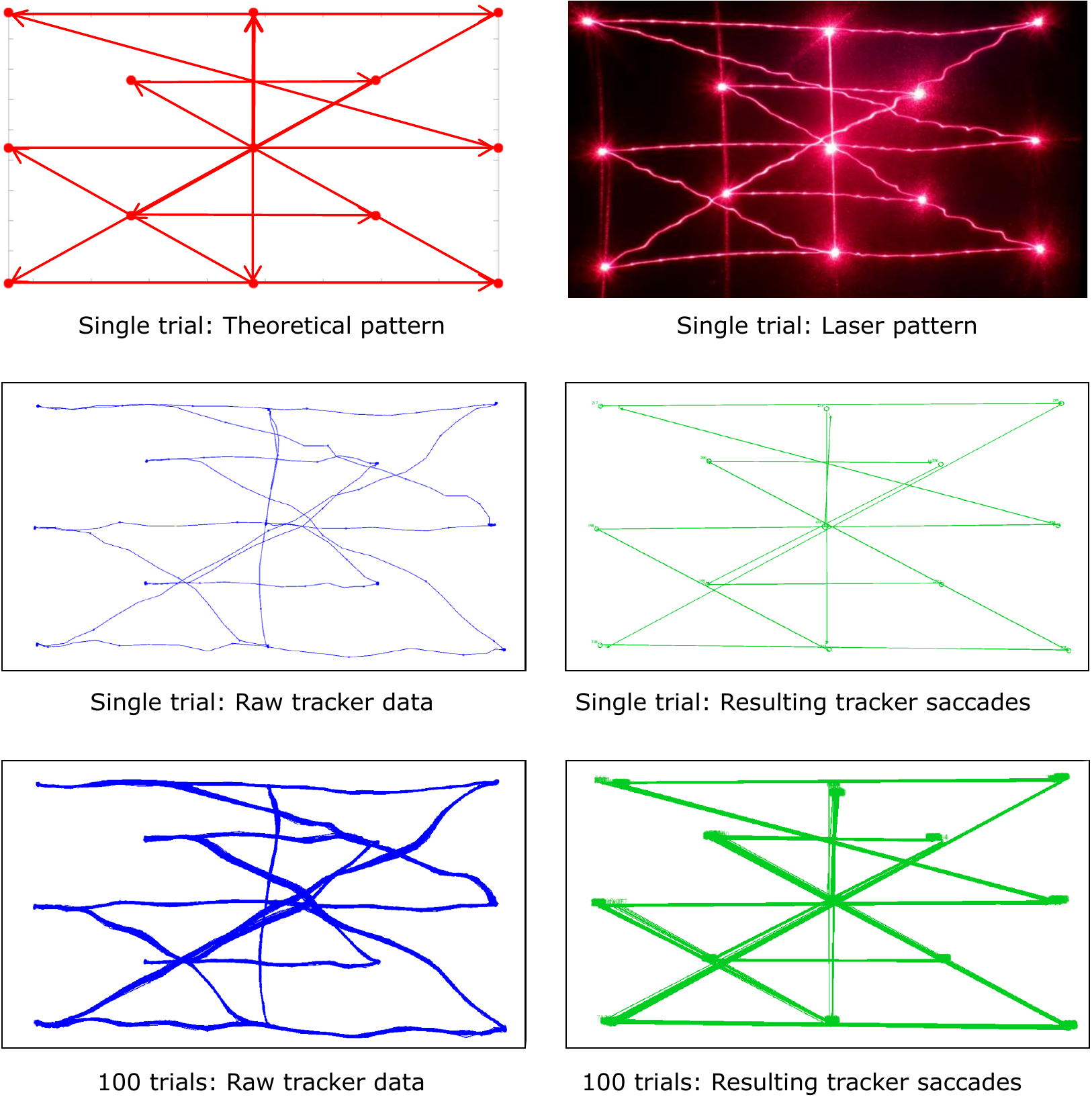}
{Eye tracker data for 13 point calibration pattern including theoretical, laser, raw and calculated saccade values. \label{fig:13point}}

Figure \ref{fig:13point} compares exemplary the planned and real laser 13-point calibration grid pattern (upper panel), the raw data and the calculated saccades for it's single trials (middle panel) and the overlays of 100 trials for the raw and calculated saccades (lower panel). The accumulated overlays for the horizontal and vertical lines look similar in quality and can be seen in the supplemental material \cite{Strazdas_Felsberg_2022}. One can intuitively draw conclusions on the precision from the similarity of the accumulated data, as the shape does not differ much from those of the single trials. 

We calculated the spatial precision of the eye tracker and RELAY for the 13-point calibration pattern, using the standard deviations (SD) to the means from the raw data's X- and Y-positions over all 100 trials (during fixation periods). They were also converted into mm and degrees of visual angle, delivering a metric for the precision of the combined eye tracker and robot system. The standard deviation values for the precision ranged from 0.3~px to 2.2~px (or 0.005° to 0.04°) for the X-axis and from 0.7~px to 2.6~px (or 0.01° to 0.04°) for Y-axis. 
Table \ref{tbl:precision} shows the values for all 13 coordinates. The calculation of accuracy and precision was computed in Microsoft Excel \cite{msexcel}. 

\begin{table}
  \caption{Precision for the 13-point calibration grid.}
  \label{tbl:precision}
  \includegraphics[width=0.999\linewidth]{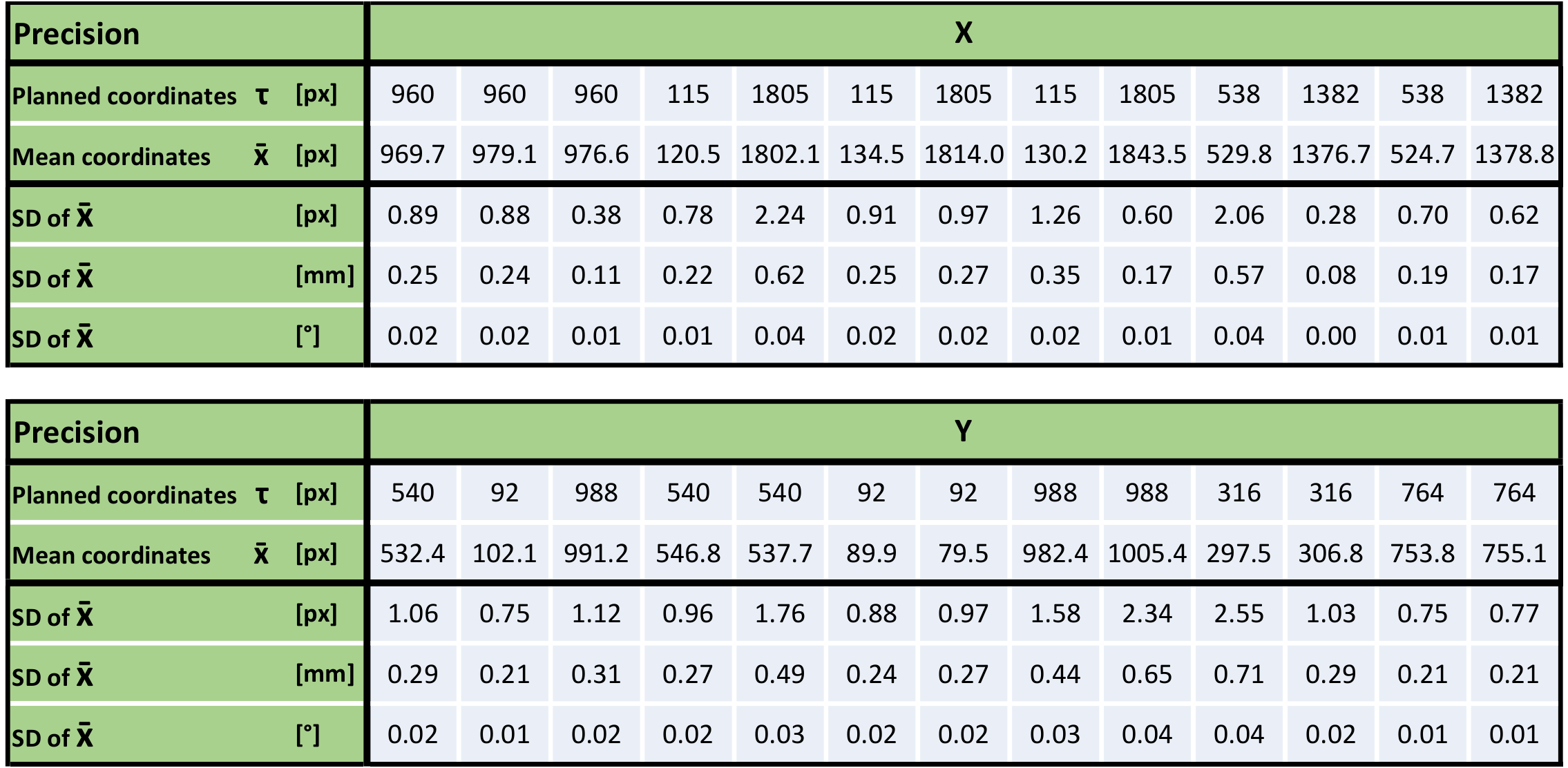}
\end{table}

\subsection{Artificial Saccades Experiment}
A total of 120,000 saccades should have been recorded (200 trials for each of the 200 artificial participants) but 2 "participants" had to be excluded from the analysis because the data set however was empty. The  saccades were aimed to lie between $4.5\degree$ and $14.1\degree$, where the human main sequence has a linear shape \cite{Bahill1975}. If one adds the acknowledged tolerance of $0.5\degree$ by \cite{SR_manual}, the range becomes $4\degree$ - $14.6\degree$. When changing between the last gaze fixation of one "participant" and the first of the next, a saccade was recorded and, depending on the position of the coordinates, lead to an amplitude which exceeded the intended size. Those were excluded from the analysis. 
Another noticeable issue was that the tracker occasionally lost track of the eye (possibly due to a reflection or the end of trackable range) during saccades that had approached the upper right corner. This caused an abort of the measured saccade and split it into two smaller ones. Those were also excluded. The described exclusion criteria resulted in a rest of 115\,681 saccades ($96.9~\%$). 

First we plotted a main sequence for all "participants" and all trials over all conditions, to check if they look similar to ones, that humans would produce and to get a rough overview. A linear regression revealed an estimated peak velocity of ($357.9\degree/s$) corresponding to a saccade of a 10° amplitude. We chose this amplitude, in order to have comparable saccades between the conditions and "participants". Indeed, it looked similar to a main sequence that would have been produced by real human participants. But due to the enormous amount of data points and overlapping coordinates it gave no further information and looked "overloaded", why we decided to include it in the appendix \cite{Strazdas_Felsberg_2022}. Then, we plotted the main sequences over all participants, for each condition separately, and performed a linear regression to get the estimated peak velocity values for an amplitude of $10\degree$. They looked very similar and so were their estimated peak velocities for dark, medium and light: $357.908\degree/s$, $357,911\degree/s$ and $357.900\degree/s$.

\Figure[!h]()[width=0.999\linewidth]{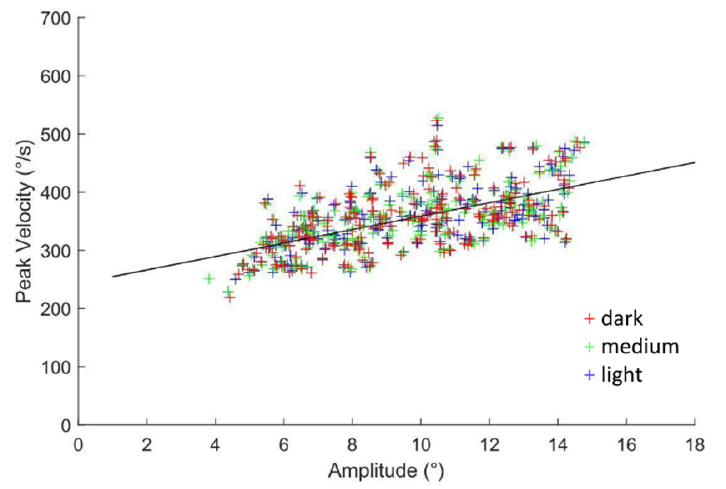}
{Main sequence for "participant 15" over all conditions. \label{fig:15buntorg}}

Figure \ref{fig:15buntorg} depicts a main sequence for a random chosen "participant" for all three brightness conditions. It becomes evident that the data points over the three conditions lie close to each other, as they should, because every condition used the same coordinates (respective amplitudes) and thus lead to similar peak velocities. The main sequences for all other "participants" were plotted as well and looked similar. They can be found as an animated graphics interchange format (GIF) in the appendix as well \cite{Strazdas_Felsberg_2022}. The plots indicate that the brightness conditions seem to have no influence on the measured peak velocities.

To test for this assumption, the regressed peak velocity values were entered in a repeated measurements analysis of variance (rm ANOVA) with the factor brightness which contained the three levels dark, medium and light. 
The rm ANOVA confirmed that there is indeed no difference between the brightness conditions which is indicated by the within-subjects (or between treatments) effect failing to reach significance: $F(2,394) = .015$, $p = .985$, partial $\eta^{2} = .000$. The resulting mean peak velocities for all brightness conditions can be observed in Figure \ref{fig:meanpeak}. 

\Figure[!h]()[width=0.999\linewidth]{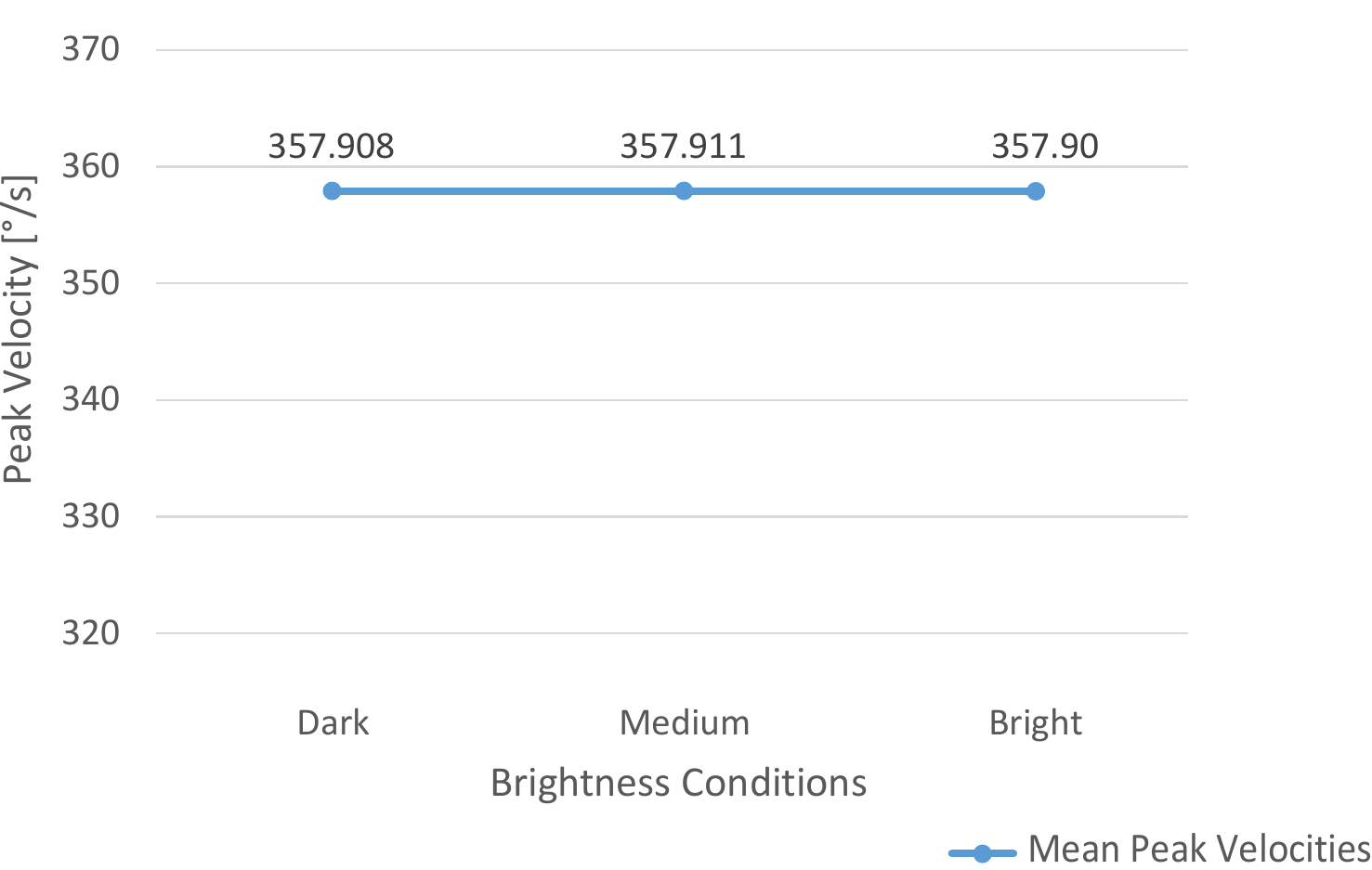}
{Mean peak velocities from the rm ANOVA over the three brightness conditions. \label{fig:meanpeak}}

Additionally, it was tested whether the recorded pupil sizes differed significantly in dependence on the brightness condition. For further evaluation, the measured values were also entered in a rm ANOVA with the same factor and levels. A Mauchly test showed a violation of sphericity, therefore the Huynh-Feldt adjustment was used for correction. With $F(1.993,74423.569) = .115$, $p = .891$ and partial $\eta^{2} = .000$, the within-subjects effect again failed to reach significance, which shows that the pupil sizes did not differ for the brightness conditions. This can also be observed in Figure \ref{fig:mean pupils bright}. 
\Figure[!b]()[width=0.999\linewidth]{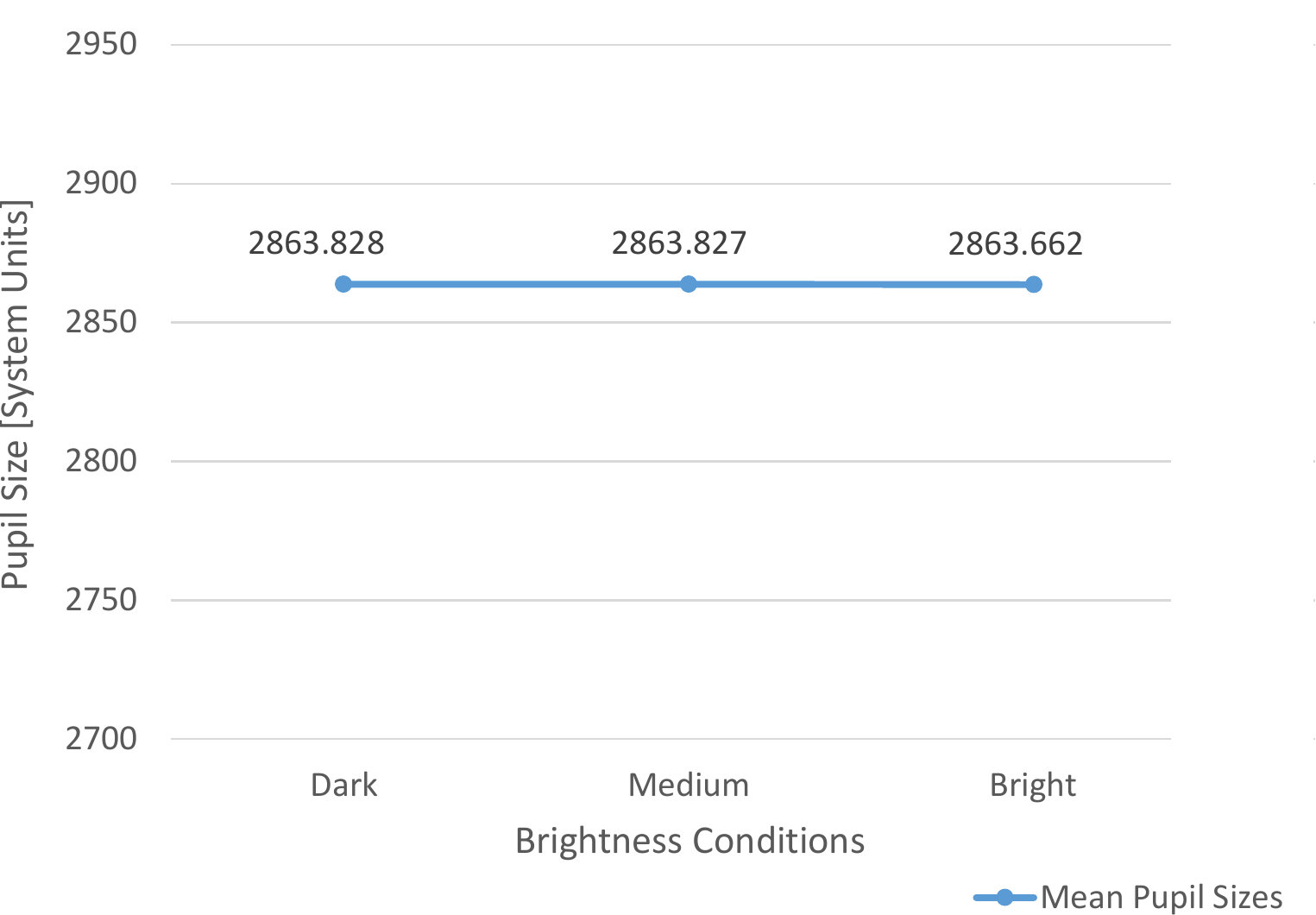}
{Mean pupil sizes from the rm ANOVA over the three brightness conditions, in the eye tracker's system units. \label{fig:mean pupils bright}}

The initial hypothesis can be stated as: If there is a brightness related artifact caused by the eye tracker, it should result in different measured peak velocities and pupil sizes across the three different brightness conditions given the same predefined amplitudes and velocities. Since all brightness conditions lead to the same measured peak velocities and pupil sizes, it can be assumed that the brightness of the environment has no influence on the peak velocity nor on the pupil size. 

As for the question, if the gaze direction has any influence on the measured pupil size, the pupil data from the precision measurement for the outer 9 coordinates from the 100 trials of the 13-point calibration grid was analyzed. This was eligible as the coordinates were precisely known so that it was possible to group the pupil size data in regard to the coordinates into directions. The resulting direction groups were: horizontal (left, middle and right) and vertical (up, middle and down). 
The values were entered into two separate rm ANOVAs with one factor (horizontal or vertical) and three levels for each direction. In order to avoid using the same values twice, the trials were split into two trial groups. The vertical analysis got the first 50 trials and the horizontal analysis got the last 50 trials. This was necessary to prevent an artificially too high correlation between the values because each of the used coordinates could have been counted as horizontal and vertical, depending on the point of view. For instance, the upper left coordinate (115,92) could technically correspond to the factor vertical as part of the upper coordinates and at the same time to the factor horizontal as part of the left coordinates. The splitting of the measured coordinates into the two groups ensured that this was not the fact. In Figure \ref{fig:mean pupils gaze} the mean pupil sizes in dependency of the gaze directions can be seen.

\Figure[!tb]()[width=0.999\linewidth]{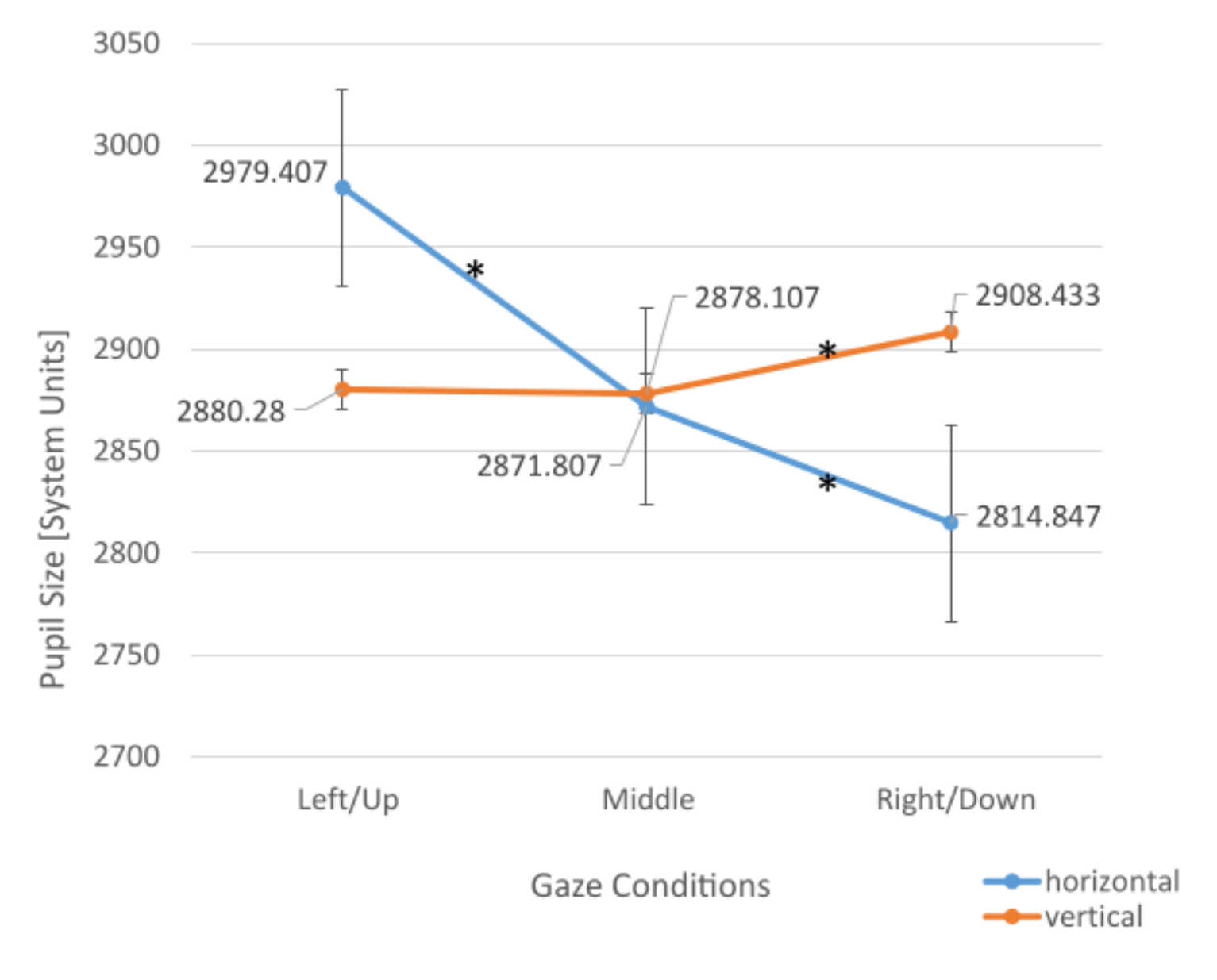}
{Mean pupil sizes from the rm ANOVAs for the gaze directions. Error bars indicate standard error. Asterisks indicate significant mean differences. \label{fig:mean pupils gaze}}

A Mauchly test showed a violation of sphericity for the horizontal gaze directions, so a Greenhouse-Geisser adjustment had to be applied. This resulted in the following statistics: $F(1.022,152.242) = 1993.581$; $p < .001$; partial $\eta^{2} = .930$. Post-hoc, pairwise and Bonferroni-adjusted comparisons revealed that the measured pupil size was significantly different for each of the possible gaze groups (left and middle, middle and right, left and right). Gazes towards the left led to the biggest pupil sizes and towards the right led to the smallest, leaving the middle between them. 

For the vertical gaze directions, the Mauchly test showed a violation of the sphericity again, which is why a Greenhouse-Geisser adjustment was applied, returning the following statistics: $F(1.030,153.475) = 86.802$; $p < .001$; partial $\eta^{2} = .368$. Post-Hoc, pairwise and Bonferroni-adjusted comparisons showed that the measured pupil sizes for gazes towards the lower coordinates differed significantly from those towards the middle or upper ones ($p < .001$). The downward gazes had the biggest measured pupil sizes, followed by upwards and the middle.  

These results show that there seems to be a bias for the measured pupil sizes depending on the driven coordinates and therefore the gaze positions.

The analysis was performed in Microsoft Excel 2013 \cite{msexcel}, Matlab R2020b and IBM SPSS Statistics 21 \cite{spss}.

\section{Conclusion and Discussion}\label{sec:conclusion}
This work had two purposes: the first was, to evaluate the accuracy and precision of RELAY and the Eyelink 1000. The results from the laser and calibration patterns (see Figure \ref{fig:13point}) show an overall superb precision and accuracy for the RELAY apparatus as well as the EyeLink 1000 tracker. The theoretical patterns were correctly represented in both the long term laser pattern exposures and the resulting saccades calculated by the eye tracker. These results indicate a high suitability for RELAY as a device for testing the EyeLink 1000 in particular and methods for video based eye tracking in general.

The differences found for the pupil sizes in dependency on the gaze positions are partly in accordance to the findings of others, like \cite{gagl2011systematic} or \cite{hayes2015}. Although they are statistically significant for some directions, it is noteworthy that if one calculates the percentage deviation from the mean ($2888.94$, in the tracker's system units), the deviations range from $-2.565\%$ (right) to $+3.131\%$ (left) for the horizontal direction and from $-0.375\%$ (middle) to $+0.675\%$ (down) for the vertical direction. Assuming that the mean reflects a pupil size of 5 mm (the size of our artificial pupil), a deviation of said percentages yields a range of $0.285~mm$ horizontally and $0.053~mm$ vertically, which is less than the thickness of a folded letter. The observed differences in this study are a lot smaller than the ones found by \cite{gagl2011systematic} or \cite{hayes2015}, who reported changes from $+5\%$ to $-13\%$, respective up to $14.4\%$, with an artificial eye. 

A further noteworthy point is, that the greatest measured values are for the left and downward directions. It can be assumed that, given the position of the eye tracker's camera in regard to the artificial eye, which was slightly left and down from it, the biggest pupil sizes occurred when the pupil was looking directly into the camera, hence appeared more round than if the gaze fell into another direction, where pupil appearance was more elliptic. This explains the changes for the horizontal positions in accordance with \cite{gagl2011systematic} or \cite{hayes2015}. The only thing that this assumption can't explain is why the pupil sizes increased slightly again from middle to upward direction. But this change was marginal ($-0.3\%$ for middle and $-0.375\%$ for upward, from the mean), so it is not clear, whether this deviation is negligible or subject to measurement uncertainty. 
For future works it could be interesting, to further investigate the geometric relationship between the eye, the tracker and the gaze direction with a tower mount in order to either find a setup with the smallest possible deviations or to develop a correction algorithm, like \cite{gagl2011systematic} and \cite{hayes2015} did.  
If a correction algorithm for the artificial eye is found, one could work on a correction with human participants. 

Although there were small differences in the measured pupil sizes, the mean peak velocities did not change. Thus, it is questionable, how big the influence of gaze position really is. A greater problem for estimating could be the distortion of the pupil that happens during saccades with real eyes and leads to the post-saccadic wobble. This is a factor which could not be tested for with the used rigid eye in this approach, but would be worth testing. 
The relationship of pupil size and gaze position seems to be bidirectional, as the pupil size is also able to influence the measured gaze position \cite{nystrom2016pupil, CHOE201648}. In this study there was only one possible influence direction because of the fixed pupil. It would be of interest to test with a variable pupil at differing and unchanged positions, in behalf to find out if both influences have the same weight. 

The second purpose of this work was to find out, if the illumination of the surroundings has an influence in the measurement of saccadic peak velocities and pupil sizes, when the latter is fixed, or so to say, if the eye tracker produces an artifact due to different brightness conditions. The presented results show that this does not seem to be the case. Therefore, it is possible that the \enquote{roof shaped} pattern for the peak velocities found by \cite{Felberg2018TheEO} is due to some yet unknown human factor, as the linear increase of pupil size with the decrease of brightness can not explain the non-linear course of peak velocities. Yet, it still needs to be tested how much the gaze direction contributed to their measured peak velocities. To address this issue, it could be advisable to let a higher number of subjects and RELAY perform the exact same saccades and compare them (preferably of fixed amplitude intervals like $\pm5\degree$, $\pm10\degree$,$\pm15\degree$ and $\pm20\degree$). The idea behind that is simple: changes in human pupil sizes and peak velocities  minus changes in robotic pupil sizes and peak velocities = rest of human factor. The fixed sizes would provide enough direct material for comparison so that one wouldn't have to estimate values via regression. This would be practicable to detect even small over- and undershoots that could be due to the gaze positions. 
As the original study by \cite{Felberg2018TheEO} had 26 participants, it would be an option, to repeat the study with more subjects and better controlled gaze directions, in order to further investigate the effect between different brightness conditions and saccade peak velocities for constant contrast conditions.

The MATLAB code and additional result presentation can be found in the appendix repository \cite{Strazdas_Felsberg_2022}.

\bibliographystyle{IEEEtran}
\bibliography{sources}





\EOD

\end{document}